\documentclass[runningheads]{llncs}
\usepackage[T1]{fontenc}
\usepackage{graphicx}
\usepackage{color}
\usepackage{multirow}
\usepackage{amsmath}
\usepackage{amsfonts}
\usepackage[misc]{ifsym}
\DeclareMathOperator*{\argmax}{argmax}
\begin{document}
\title{Online Easy Example Mining for Weakly-supervised Gland Segmentation from Histology Images}
\author{Yi Li\inst{1,\thanks{Co-first authors}} \and Yiduo Yu\inst{1,^{\star}} \and Yiwen Zou\inst{1,^{\star}} \and
Tianqi Xiang\inst{1} \and
Xiaomeng Li\inst{1,2(\textrm{\Letter})}}
\titlerunning{Online Easy Example Mining for Weakly-supervised Gland Segmentation}
\authorrunning{Yi Li et al.}
%
\institute{Department of Electronic and Computer Engineering, The Hong Kong University of Science and Technology, Hong Kong, China \\ \email{eexmli@ust.hk}\\ \and
HKUST Shenzhen Research Institute, Shenzhen, China}
\maketitle              
\begin{abstract}
Developing an AI-assisted gland segmentation method from histology images is critical for automatic cancer diagnosis and prognosis; however, the high cost of pixel-level annotations hinders its applications to broader diseases. Existing weakly-supervised semantic segmentation methods in computer vision achieve degenerative results for gland segmentation, since the characteristics and problems of glandular datasets are different from general object datasets. We observe that, unlike natural images, the key problem with histology images is the confusion of classes owning to morphological homogeneity and low color contrast among different tissues. To this end, we propose a novel method \emph{Online Easy Example Mining} (OEEM) that encourages the network to focus on credible supervision signals rather than noisy signals, therefore mitigating the influence of inevitable false predictions in pseudo-masks. According to the characteristics of glandular datasets, we design a strong framework for gland segmentation. Our results exceed many fully-supervised methods and weakly-supervised methods for gland segmentation over 4.4\% and 6.04\% at mIoU, respectively. Code is available at https://github.com/xmed-lab/OEEM.

\end{abstract}

\section{Introduction}

Accurate gland segmentation is one crucial prerequisite step to obtain reliable morphological statistics that indicate the aggressiveness of tumors. With the advent of deep learning and whole slide imaging, considerable efforts have been devoted to developing automatic semantic segmentation algorithms from histology images~\cite{srinidhi2021deep}. These methods require massive training data with pixel-wise annotations from expert pathologists~\cite{chen2017deeplab,ronneberger2015u,zhao2017pyramid}. However, obtaining pixel-wise annotation for histology images is expensive and labor-intensive. 
To reduce the annotation cost, designing a weakly-supervised segmentation method that only requires a patch-level label is highly desirable.

\begin{figure}
\begin{centering}
\includegraphics[scale=0.45]{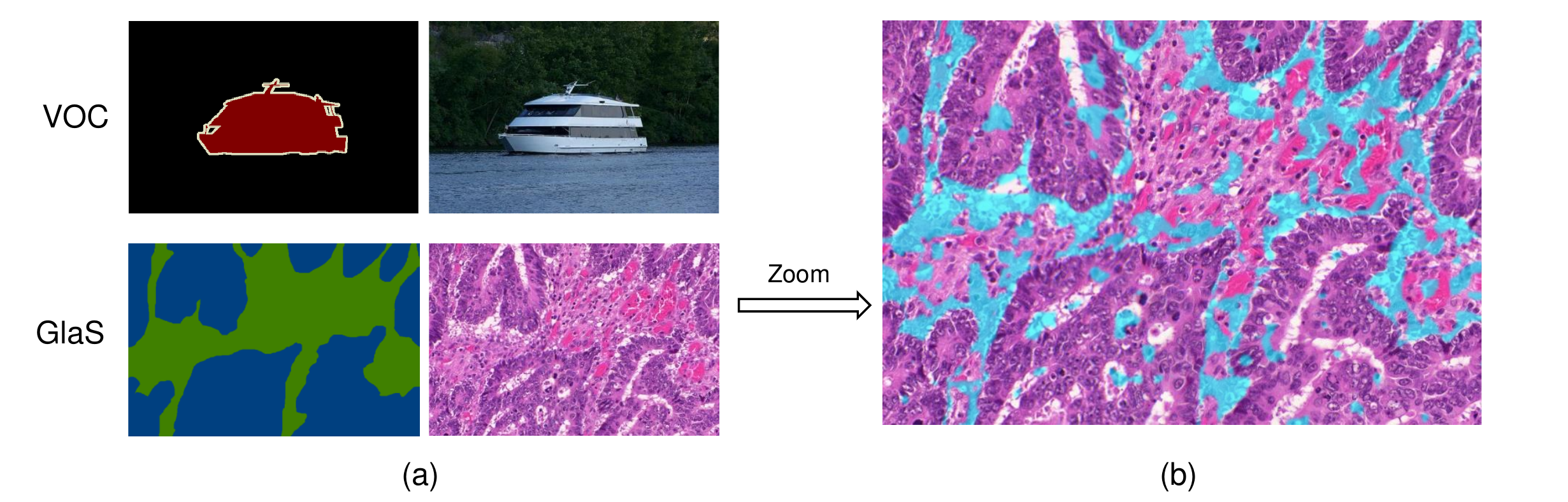}
\par\end{centering}
\caption{\label{dataset} (a): Examples of VOC (general dataset) and GlaS (glandular dataset). The \textcolor{blue}{blue} region refers to glandular tissues, and the \textcolor{green}{green} region refers to non-glandular tissues. The characteristics of GlaS are morphological homogeneity, obvious overlaps, and low color contrast among different tissues. (b): Confusion regions (highlighted in \textcolor{cyan}{cyan}) in weakly-supervised gland segmentation.}
\end{figure}

To our best knowledge, there are no prior studies for weakly-supervised gland segmentation from histology images. For other medical datasets, weakly-supervised segmentation methods are mainly based on multiple instance learning (MIL) \cite{xu2014deep,jia2017constrained,xu2019camel}, which requires at least two types of image-level labels to train the classifier. 

However, this method is not applicable to our task since all our training images contain glandular tissues, \emph{i.e.}, we only have one type of image-level label. Another limitation of MIL is the low quality of the pseudo-mask, which is block-like and coarse-grained because MIL regards all pixels within one patch as the same class.
For general weakly-supervised semantic segmentation (WSSS) approaches in computer vision, the pseudo-mask is more fine-grained with pixel-level prediction via CAM \cite{zhou2016learning}. Nevertheless, algorithms in general WSSS~\cite{kolesnikov2016seed,ahn2018learning,zeng2019joint,chang2020weakly,wang2020self} do not suit glandular datasets because its core problem is \emph{local activation} resulting from different representations in one object. While for glandular datasets, the key problem is \emph{confusion among classes}, owning to the morphological homogeneity, obvious overlap, and low color contrast of targets.

As shown in Fig.\ref{dataset} (a), compared to the natural images with apparent color differences, our gland images have a similar color distribution and morphological affinity between different tissues.
For these confusing regions in Fig.\ref{dataset} (b), techniques in general object datasets like saliency detection~\cite{zeng2019joint} and affinity learning~\cite{ahn2018learning} are invalid because these methods require \emph{salient differences} between targets and background.

To solve the confusion problem and avoid the above issues, our key idea is that the network should highlight the training with credible supervision and down-weight the learning with noisy signals. To this end, we propose the Online Easy Example Mining (OEEM) to distinguish easy and confusing examples in the optimization. Specifically, we develop a metric based on the normalized loss to achieve this goal, where pixels with lower losses are dynamically assigned higher weights in a batch.

Notably, our method is different from the existing online hard example mining methods~\cite{shrivastava2016training,lin2017focal}. For instance, online hard example mining \cite{shrivastava2016training} and focal loss \cite{lin2017focal} in object detection are hard to transfer to weakly supervised scenarios, since they amplify the noise in pseudo-mask. 

Moreover, we design a powerful framework for weakly-supervised gland segmentation and report its fully-supervised result of the segmentation stage. With this strong baseline, our method excels many previous fully supervised methods~\cite{valanarasu2021medical,valanarasu2021kiu,wang2021uctransnet} on the GlaS~\cite{sirinukunwattana2017gland} dataset, notably outperforming the prior best  method~\cite{wang2021uctransnet} by 4.6\% on mIoU.  
Importantly, even with such a high backbone, our proposed OEEM further increases the performance by around 2.0\% mIoU in weakly settings. And our weakly-supervised result surpasses the influential and general WSSS methods \cite{Wang_2020_CVPR,lee2021anti,chang2020weakly} by over 6.04\% at mIoU. 

The main contributions of our work are summarized as follows:
\begin{itemize}
    \item We point out that the key problem of gland segmentation is the confusion caused by the homogeneity of histology images, rather than the local activation problem that most WSSS methods in computer vision try to solve. 
    
    \item We propose the Online Easy Example Mining (OEEM) to mine the credible supervision signals in pseudo-mask with proposed normalized loss, thus mitigating the damage of confused supervisions for gland segmentation.
    
    \item 
    We design a strong framework for gland segmentation. Our fully-supervised and proposed weakly-supervised OEEM surpasses the existing fully- and weakly-supervised methods for gland segmentation, respectively. 

\end{itemize}

\section{Method}

\subsection{Overall Framework}
This part introduces the overall pipeline of weakly-supervised gland segmentation, consisting of two stages. As shown in Fig.~\ref{framwork}, our framework starts from the classification stage, which is trained using patch-level supervision only. Then we synthesize the pseudo pixel-level mask based on CAM~\cite{zhou2016learning} for the training set. The pseudo-masks with original images are then passed to the segmentation stage in the manner of fully-supervised segmentation. To reduce the adverse impacts resulting from the noise in pseudo-masks, we adopt the proposed OEEM during the optimization of the segmentation model. Note that only the segmentation network is used to generate final predictions. The details of these two stages are shown below:

\begin{figure}
\begin{centering}
\includegraphics[scale=0.5]{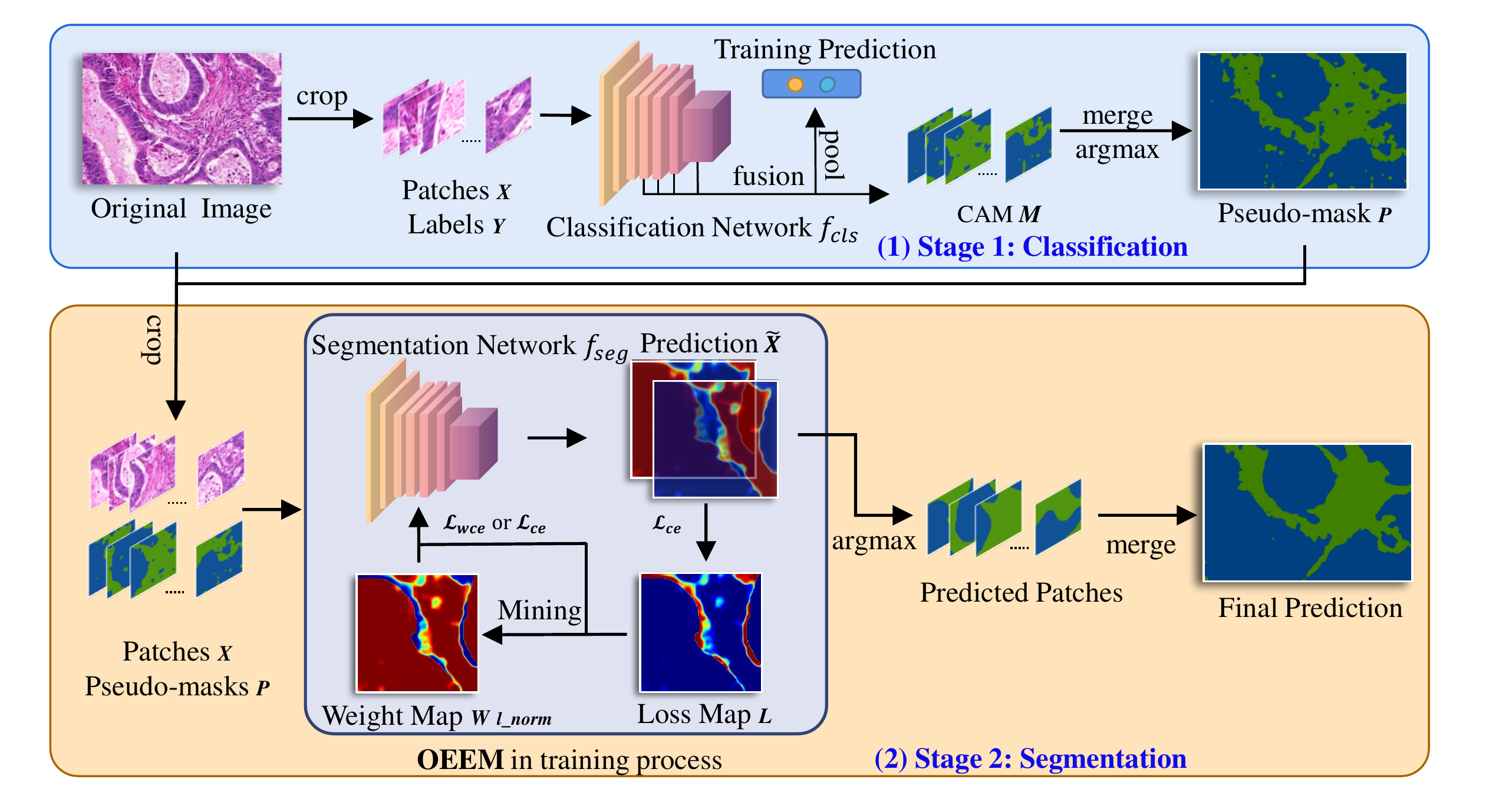}
\par\end{centering}
    \caption{\label{framwork} Overview of our proposed weakly-supervised gland segmentation method with OEEM. (a) Classification pipeline for pseudo-mask generation from CAM. (b) Segmentation pipeline. We use weighted cross-entropy loss $\mathcal{L}_{wce}$ with weight map $\boldsymbol{W}_{l\_norm}$ for multi-label patches. Here, the glandular tissues are shown in \textcolor{blue}{blue} and non-glandular tissues are shown in \textcolor{green}{green}.}
\end{figure}

\textbf{Classification:}
We denote the input image as $\boldsymbol{X} \in \mathbb{R}^{C \times H \times W}$, and its patch-level label as $\boldsymbol{Y}$, where $\boldsymbol{Y}$ is a one-hot vector of $[y_{1}, y_{2}, ..., y_{n}]$ and $n$ is the number of classes in the dataset. $feat = f_{cls}(X, \phi_{feat})$ is the predicted feature map via network $f_{cls}$ and its parameters $\phi_{feat}$ except that of classifier. Note that feature maps of last three stages are fused with interpolations before the classifier for better representation. Then we get the CAM $\boldsymbol{M} \in \mathbb{R}^{n \times H \times W}$ by multiply the weights of classifier $\phi_{cls}$ to $feat$ without average pooling as:
\begin{equation}
    \boldsymbol{M} = f_{cls}(X, \phi_{feat}) \cdot \phi_{cls}
\end{equation}

The idea of CAM is that the feature maps of CNN architecture contain spatial information of the activated regions in the image, where the classification model is paying attention to. 
We apply argmax operation on $M$ along the category dimension and get a 2-dim pseudo-mask $\boldsymbol{P} \in \mathbb{R}^{H \times W}$ as Eq.\ref{pseudo}. $P$ is subsequently sent to segmentation stage.
\begin{equation}
    \boldsymbol{P}=\argmax(\boldsymbol{M}_{:,h,w}) ,\forall h,w
\label{pseudo}
\end{equation}

\textbf{Segmentation:}
We now train the segmentation model in a fully-supervised manner using pseudo-mask $P$. Denoting the segmentation network as $f_{seg}$ with parameters $\phi_{seg}$, the eventual prediction result $\boldsymbol{\tilde{X}}$ is derived as $\boldsymbol{\tilde{X}} = f_{seg}(\boldsymbol{X},\phi_{seg})$. Due to the inevitable noisy responses in pseudo-mask, the supervision $\boldsymbol{P}$ introduces many confusing signals to the optimization of segmentation. To cope with this issue, we need to direct supervised learning by weakening the guidance in the confusing region. Based on this motivation, we propose the OEEM to reweight the loss map for better usage of credible and clean supervision.

\subsection{Online Easy Example Mining}
Many solutions are proposed in general object-based weakly-supervised segmentation, such as seed expansion\cite{kolesnikov2016seed}, self-attention \cite{Wang_2020_CVPR}, sub-category clustering \cite{chang2020weakly} and affinity \cite{ahn2018learning}. But due to the changed core problem, we are concerned about how to deal with these confusing noises. Previous works apply offline loss weight based on uncertainty \cite{li2021uncertainty}, or online loss clip relying on tuning hyperparameters \cite{li2021pseudo}. Besides, some sampling methods like OHEM \cite{shrivastava2016training}, and focal loss \cite{lin2017focal} improve the training efficiency via mining the hard examples in fully-supervised segmentation. But the noises will be mined together with hard examples in our case.

Unlike the above methods, we propose the online easy example mining to learn from the credible supervision signals dynamically. Specifically, our OEEM modifies the standard cross-entropy loss $\mathcal{L}_{ce}$ by multiplying a loss weight via a metric that indicates the difficulty. 

Denote our segmentation prediction map $\boldsymbol{\tilde{X}}$ and ground truth map $\boldsymbol{Y}$. We realize the reweighting scheme by the weighted cross-entropy loss as:
\begin{equation}
    \mathcal{L}_{wce} =-\sum_{j=0}^{w}\sum_{i=0}^{h} \boldsymbol{W}_{i,j}\cdot \log \frac{{\rm exp}({\boldsymbol{\tilde{X}}_{\boldsymbol{Y}_{i,j},i,j}})}{\sum_{k=1}^C ({\rm exp}{\boldsymbol{\tilde{X}}_{k, i,j}})} \cdot 1 
\label{wce}
\end{equation}
where $\boldsymbol{W} \in \mathbb{R}^{H\times W}$ is the loss weight. To get this loss weight, we pick the loss scattered on loss map $\boldsymbol{L} \in \mathbb{R}^{H\times W}$ and confidence on predicted score map as metrics to indicate the learning difficulty. And base on these metrics, we propose four types of loss weight maps to mine the easy examples.

The first two weight types are based on the metric of confidence. The motivation of Eq.\ref{w_max} is that easy examples are usually of high confidence. We firstly apply softmax operation $sm$ on the category dimension to normalize the score to $[0, 1]$, and select the maximum value as metric to form the weight map $\boldsymbol{W}_{c\_max}$. The second type Eq.\ref{w_diff} uses the difference of confidence as the metric, since comparable confidences indicate harder examples.
\begin{equation}
    \boldsymbol{W}_{c\_max} = \max((\rm{sm}(\boldsymbol{\tilde{X}}_{:,h,w}))), \forall h,w
\label{w_max}
\end{equation}
\begin{equation}
    \boldsymbol{W}_{c\_diff} = \max((\rm{sm}(\boldsymbol{\tilde{X}}_{:,h,w}))) - \min((\rm{sm}(\boldsymbol{\tilde{X}}_{:,h,w}))), \forall h,w
\label{w_diff}
\end{equation}
The other two types are based on the loss value. Different from the confidence, the loss is obtained from both confidence and pseudo-ground-truth with more information. The noises with high confidences on the false category have high loss values, and those pixels supervised by clean labels have lower loss values. To get higher loss weight for easy examples, we apply a minus sign to the loss map $\boldsymbol{L}$ and deploy the softmax function $sm$ on the $hw$ dimension with a division of its mean value. We name this process to normalized loss as Eq.\ref{w_loss}. At last, we combine max confidence and normalized loss as Eq.\ref{w_mix}.
\begin{equation}
    \boldsymbol{W}_{l\_norm} = \frac{\rm{sm}(- \boldsymbol{L})}{\rm{mean}(\rm{sm}(- \boldsymbol{L}))}
\label{w_loss}
\end{equation}
    
\begin{equation}
    \boldsymbol{W}_{lc\_mix} = \frac{\rm{sm}(- \boldsymbol{L})}{\rm{mean}(\rm{sm}(- \boldsymbol{L}))} \cdot \rm{max}((\rm{sm}(\boldsymbol{\tilde{X}}_{:,h,w}))), \forall h,w
\label{w_mix}
\end{equation}

Empirical experience suggests that $\boldsymbol{W}_{l\_norm}$ in Eq.\ref{w_loss} performs best. So we select it as the metric of reweighting. And note that some images only contain one type of artifact that should not yield any noise. Hence, we use original cross-entropy loss $\mathcal{L}_{ce}$ without OEEM for images with number of classes $n=1$. The final loss of segmentation is then expressed as:
\begin{equation}
\mathcal{L} =
\begin{cases}
\mathcal{L}_{wce} & n > 1, s.t. \boldsymbol{W}= \boldsymbol{W}_{l\_norm} \\
\mathcal{L}_{ce} & n = 1. 
\end{cases}
\end{equation}
\subsection{Network training}
Our model is implemented with PyTorch and is trained with one NVIDIA GeForce RTX 3090 card. For the classification part, we adopt ResNet38~\cite{wu2019wider} as the backbone. We use an SGD optimizer with a polynomial decay policy at a learning rate of 0.01. The batch size is 20, and the model is trained for 20 epochs. Data augmentation includes random flip, random resized crop, and all patches are normalized by the calculated mean and variance of the GlaS dataset. We also utilize the multi-scale test with scales of [1, 1.25, 1.5, 1.75, 2] for CAM generation. For the segmentation part, we use PSPNet~\cite{zhao2017pyramid} with backbone ResNet38. The optimizer is SGD in poly scheduler at learning rate $5e - 3$, the batch size is 32, and the model is trained for 10000 iterations. Data augmentation includes random flip, random crop, and distortion. In the inference process, we apply a multi-scale test with scales of [0.75, 1, 1.25, 1.5, 1.75, 2, 2.5, 3] at crop size 320 and stride 256 for robust results. Note that all the weakly-supervised methods in Tab.\ref{compare sota} deploy the same settings for fair comparisons.

\section{Experiments}

\subsection{Dataset}

We carry out experiments on the Gland Segmentation in Histology Images Challenge (GlaS) Dataset~\cite{sirinukunwattana2017gland}. It contains 165 images derived from 16 Hematoxylin, and Eosin (H\&E) stained histological Whole Slide Images (WSIs) of stage T3 or T42 colorectal adenocarcinoma.

Following previous works~\cite{valanarasu2021medical,valanarasu2021kiu,wang2021uctransnet}, we split the data into 85 training images and 80 test images, which are separated by patients as original dataset without patch shuffle. There is no classifiable image-level label, since glands exist in each image. So we crop patches at side 112 and stride 56 to get balanced patch-level labels from masks, with a one-hot label indicating whether it contains glandular and non-glandular tissues for each patch. Note that all the patches are merged to the original sizes for evaluation in the metric of mIoU.

\subsection{Compare with State-of-the-arts}

\begin{table}
\begin{centering}
\caption{\label{compare sota} Results of gland segmentation on the GlaS dataset. ``-'' refers to not reported.}
\setlength\tabcolsep{3pt}
\begin{tabular}{cccccc}
\hline 
Method & Backbone & Supervision & mIoU (\%) & Dice (\%) & F1 (\%) \tabularnewline
\hline
{Unet~\cite{ronneberger2015u}} & Unet & fully & 65.34 & 79.04 & 77.78\tabularnewline
{Res-UNet~\cite{xiao2018weighted}} & Res-UNet & fully & 65.95 & 79.48 & 78.83\tabularnewline
{MedT~\cite{valanarasu2021medical}} & Vision Transformer & fully & 69.61 & 82.08 & 81.02\tabularnewline
{KiU-Net~\cite{valanarasu2021kiu}} & KiU-Net & fully & 71.31 & 83.25 & - \tabularnewline 
{UCTransNet~\cite{wang2021uctransnet}} & UCTransNet & fully & 82.24 & 90.25 & - \tabularnewline
{ours w/o OEEM} & {PSPNet \& ResNet38} & fully & \textbf{86.84} & \textbf{92.96} & \textbf{93.24} \tabularnewline
\hline 
{SEAM~\cite{wang2020self}} & {PSPNet \& ResNet38} & weakly & 66.11 & 79.59 & 79.50 \tabularnewline
{Adv-CAM~\cite{lee2021anti}} & {PSPNet \& ResNet38} & weakly & 68.54 & 81.33 & 81.36 \tabularnewline
{SC-CAM~\cite{chang2020weakly}} & {PSPNet \& ResNet38} & weakly & 71.52 & 83.40 & 83.32\tabularnewline
{ours w/o OEEM} & {PSPNet \& ResNet38} & weakly & 75.64 & 86.13 & 82.36 \tabularnewline
{OEEM} & {PSPNet \& ResNet38} & weakly & \textbf{77.56} & \textbf{87.36} & \textbf{87.35}\tabularnewline
\hline 
\end{tabular}
\par\end{centering}
\end{table}

\noindent \textbf{Compare with fully-supervised methods.} 
We compare the final results after the segmentation step in Tab. \ref{compare sota}. The result of our method without OEEM shows that we have already constructed a robust baseline model that excels in many fully-supervised settings like Unet~\cite{ronneberger2015u} and MedT~\cite{valanarasu2021medical}. This suggests that our classification model based on ResNet38 with a multi-test mechanism performs quite well for the pseudo-mask generation. 

\noindent \textbf{Compare with weakly-supervised methods.} 
In Tab.\ref{compare sota}, we firstly propose a powerful baseline, whose fully-supervised result is 86.84\%, significantly beyond previous state-of-the-arts. Based on this framework, our weakly-supervised result is higher than most fully-supervised methods. Even under such a high baseline, our OEEM still works fine and increases our baseline from 75.64\% to 77.56\%. We also experimented with other influential weakly-supervised methods in the general object segmentation domain, such as SEAM, SC-CAM, and Adv-CAM. There is a large margin of at least 6.04\% comparing to the proposed OEEM method. This is because the confusion owing to morphological homogeneity, low color contrast and serious overlap of tissue cells obstructs the network learning, leading to low quality pseudo-masks of classification stage. 

\begin{figure}[ht]
\begin{centering}
\includegraphics[scale=0.4]{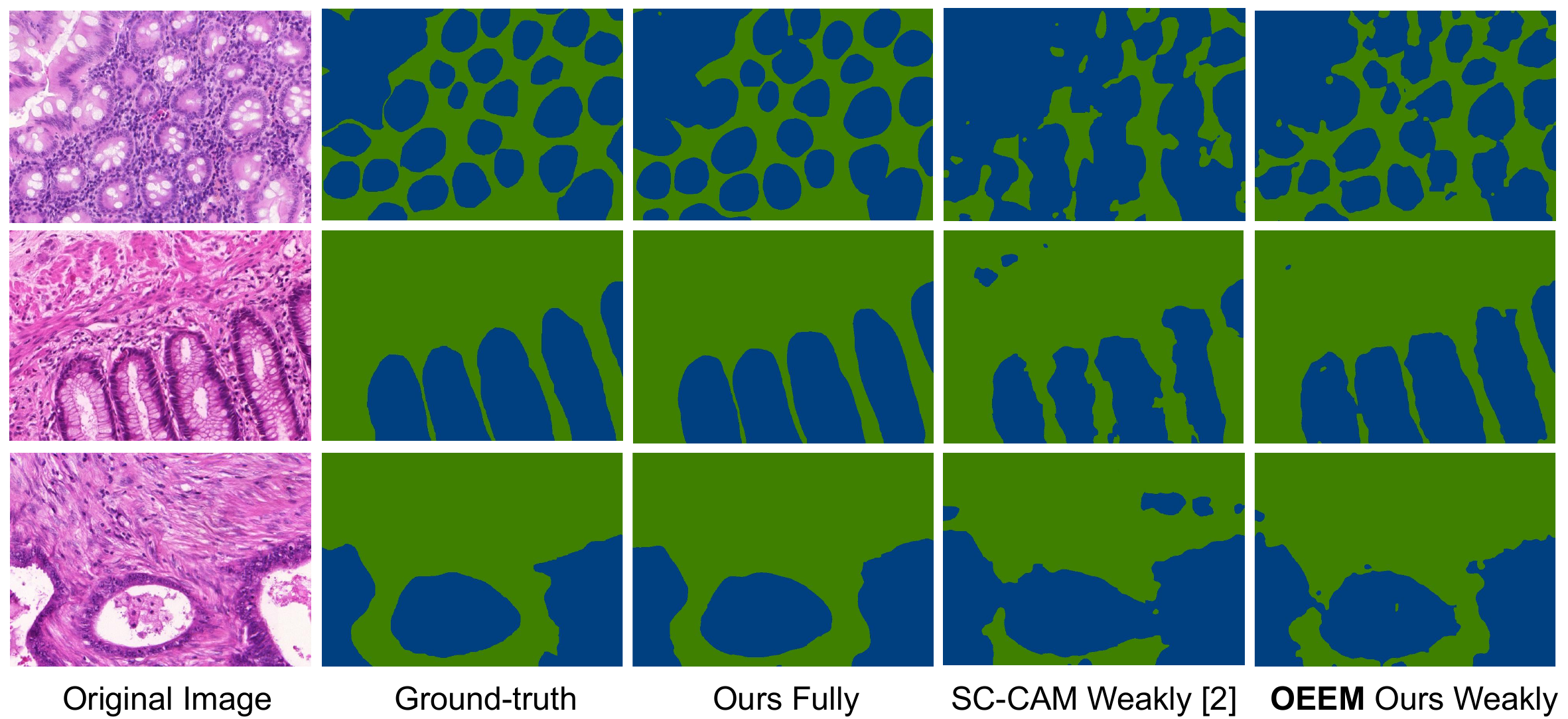}
\par\end{centering}
\caption{\label{Qualitative results} Qualitative results compared with ground-truth, fully and other weakly results. The glandular tissues are shown in \textcolor{blue}{blue}
and non-glandular tissues are shown in \textcolor{green}{green}.}
\end{figure}

\noindent \textbf{Visualization} 
Here we show some qualitative visualization results compared to SEAM~\cite{wang2020self} in Fig.\ref{Qualitative results}. The SEAM prediction appears to be coarse and inaccurate with many square-shaped regions. This suggests that SEAM is unsuitable for the GlaS dataset and the Affine Transformation design of the method further poses some difficulties in the training process. In contrast, our OEEM prediction is more accurate and smooth, which seems quite similar to the fully-supervised result.

\subsection{Ablation Study}

\noindent \textbf{Performance gains in overall framework.}
We show the performances gains in the framework by listing the results of different predictions. The SEAM is different from our framework in the classification stage, with a drop at mIoU 15.89\% owning to the shifted core problem in the histology dataset. The performance gain of our framework from CAM to pseudo-mask and segmentation prediction are 2.12\%, 9.64\% respectively. Note that OEEM shares the same pseudo-mask with our baseline and improves the performance by about 2\%.

\begin{table}

\caption{\label{results} Performance of the framework at metric mIoU (\%). CAM denotes the output of the classification stage on training set. Pseudo-mask is the refined CAM after using the patch-level labels to eliminate the non-existing tissues. Prediction is the result of the segmentation state on testing set.}
\begin{centering}
\setlength\tabcolsep{8pt}
\begin{tabular}{cccc}
\hline 
 Method & CAM & Pseudo-mask & Prediction \tabularnewline
\hline
SEAM~\cite{wang2020self} & 52.03 & 60.48 & 66.11 \tabularnewline

 {ours w/o OEEM} & \multirow{2}{*}{67.92} & \multirow{2}{*}{70.04} & 75.64 \tabularnewline
 {OEEM} & & & \textbf{77.56} \tabularnewline

\hline 
\end{tabular}
\par\end{centering}

\end{table}

\noindent \textbf{Effectiveness of OEEM with normalized loss.}
For the segmentation part, we compare four OEEM weight metrics and OHEM~\cite{shrivastava2016training} with our baseline result in Tab.~\ref{metrics}.  Unlike hard example mining in fully supervised learning \cite{shrivastava2016training}, pseudo-masks from weakly supervision exist massive noise. It means hard samples and false samples are intertwined and indistinguishable. So we see that OHEM even introduces a performance drop due to the noise included. Thus, we mine the easy samples to make the supervision credible and mitigate the influence of noise from hard samples. Among which, \textit{normalized loss} performs the best. This could be attributed to the help of pseudo ground-truth in computing the weight metrics. Additionally, normalization via softmax amplifies the loss gaps and emphasizes the clean samples more. Compared to the baseline model, our OEEM strategy improves 1.92\% in mIoU, which is essential to our pipeline.

\begin{table}
\caption{\label{metrics} Segmentation results of reweighting metrics in OEEM with baselines.}
\begin{centering}
\setlength\tabcolsep{5pt}
\begin{tabular}{c|cccccc}
\hline
Metric & Baseline & OHEM \cite{shrivastava2016training} & $\boldsymbol{W}_{c\_max}$ & $\boldsymbol{W}_{c\_diff}$ & $\boldsymbol{W}_{l\_norm}$ & $\boldsymbol{W}_{lc\_mix}$ \tabularnewline
\hline
mIoU (\%) & 75.64 & 75.49 & 75.93 & 75.43 & \textbf{77.56} & 77.19 \tabularnewline
\hline 
\end{tabular}
\par\end{centering}
\end{table}

\section{Conclusion}

This paper proposes a novel online easy example mining method for weakly-supervised gland segmentation from histology images, where only patch-level labels are provided. Our main motivation is that, unlike natural images, the key problem of histology images is the confusion among classes due to its low color contrast among different tissues, making it challenging for gland segmentation. Such a property degenerates many existing weakly-supervised methods in computer vision. Our proposed OEEM focuses on training with credible supervision and down-weight the training with noisy signals. 
Experimental results demonstrated that our method can outperform other weakly-supervised methods by a large margin.

\section*{Acknowledgement}
This work was supported by Shenzhen Municipal Central Government Guides Local Science and Technology Development Special Funded Projects under

\noindent2021Szvup139 and Foshan HKUST Projects under FSUST21-HKUST11E.

\bibliographystyle{splncs04}
\bibliography{miccai_bib}
\end{document}